\documentclass{Interspeech2024}

\usepackage{multirow}
\usepackage{hyperref}
\usepackage{pgfplots}
\pgfplotsset{compat=1.18} 

\interspeechcameraready 

\usepackage{tikz}
\usepackage{everypage}
\newcommand\BackgroundText{%
  \begin{tikzpicture}[remember picture,overlay]
    \node [rotate=0, scale=1.3, text opacity=1.0, color=gray] at (current page.south west) [anchor=south west, xshift=0.7cm, yshift=1.9cm] {accepted at Interspeech 2024};
  \end{tikzpicture}
}
\AddEverypageHook{\BackgroundText}

\title{Probing the Feasibility of Multilingual Speaker Anonymization}

\name{Sarina}{Meyer}
\name{Florian}{Lux}
\name{Ngoc Thang}{Vu}

\address{Institute for Natural Language Processing, University of Stuttgart, Germany}
\email{sarina.meyer@ims.uni-stuttgart.de}


\newcommand{\asvorig}{$\mathrm{ASV}_{\mathrm{orig}}^{\mathrm{LS}}$}
\newcommand{\asvanon}{$\mathrm{ASV}_{\mathrm{anon}}^{\mathrm{LS}}$}
\newcommand{\asvvox}{$\mathrm{ASV}_{\mathrm{orig}}^{\mathrm{Vox}}$}
\newcommand{\asrlibri}{$\mathrm{ASR}^{\mathrm{eval}}_{\mathrm{libri}}$}
\newcommand{\asrwhisper}{$\mathrm{ASR}^{\mathrm{eval}}_{\mathrm{Wh}}$}
\newcommand{\asranonwhisper}{$\mathrm{ASR}_{\mathrm{Wh}}$}
\newcommand{\asrmms}{$\mathrm{ASR}^{\mathrm{eval}}_{\mathrm{MMS}}$}
\newcommand{\asren}{$\mathrm{ASR}_{\mathrm{en}}$}

\definecolor{color1}{HTML}{00b159}
\definecolor{color2}{HTML}{f37735}
\definecolor{color3}{HTML}{00aedb}
\definecolor{color4}{HTML}{ffc425}
\definecolor{color5}{HTML}{d11141}

\begin{document}

\maketitle

\begin{abstract}
    In speaker anonymization, speech recordings are modified in a way that the identity of the speaker remains hidden. While this technology could help to protect the privacy of individuals around the globe, current research restricts this by focusing almost exclusively on English data. In this study, we extend a state-of-the-art anonymization system to nine languages by transforming language-dependent components to their multilingual counterparts. Experiments testing the robustness of the anonymized speech against privacy attacks and speech deterioration show an overall success of this system for all languages. The results suggest that speaker embeddings trained on English data can be applied across languages, and that the anonymization performance for a language is mainly affected by the quality of the speech synthesis component used for it. 
\end{abstract}

\section{Introduction}
Speech contains a lot of information about the speaker, such as their identity, age, or emotion \cite{kroeger2020privacy}. 
This poses a privacy risk to anyone using speech technology as this data might be extracted and misused without the user's knowledge or consent.
Thus, speaker anonymization aims to modify speech recordings 
such that they remain usable for a target application while hiding the identity or personal information of the speaker. However, current tools support almost only English speech and thus exclude privacy protection for billions of people.

Approaches to this task are divided into signal processing \cite{patino21speaker, dubagunta2022adjustable} and voice conversion methods \cite{fang2019speaker, shamsabadi2022differentially, miao2023speaker, panariello23vocoder}, with latter being generally more robust against privacy attacks. They convert the speech content into intermediate representations in which the speaker-specific and situation-specific (e.g., linguistic content, prosody) information is disentangled. After modifying the speaker-specific into an anonymous representation, they are synthesized back to audio. 
While most modern systems employ latent representations to encode speech, our state-of-the-art approach described in \cite{meyer2023prosody} reduces the linguistic and prosodic content to human-understandable sequences to avoid unintentional leakage of personal information in these latent spaces. This produces favorable objective results on clean English data, however, its automatic speech recognition (ASR) and text-to-speech (TTS) components introduce a strong language dependence.

So far, voice privacy research is mainly limited to English speech. 
There are few signal processing based approaches for other languages like Spanish \cite{magarinos2017reversible} and Finnish \cite{tavi2022improving}, but the studies still remain monolingual. 
The only multilingual exception is the research of Miao et al. \cite{miao2023speaker, miao2022language, miao2022analyzing} who develop a language-independent system with a language-invariant content encoder producing latent representations. However, they test it only for English and Mandarin, despite training their vocoder in \cite{miao2022analyzing} also on German, Italian and Spanish. 

\begin{figure}
    \centering
    \resizebox{\columnwidth}{!}{
    \includegraphics{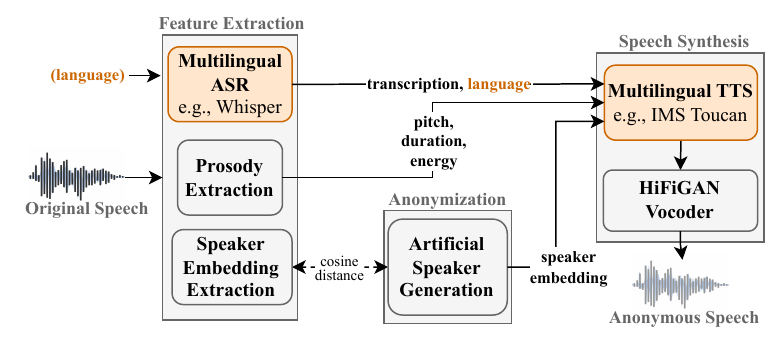}
     }
    \caption{Speaker anonymization system, adapted from \cite{meyer2023prosody}. Proposed multilingual modifications are displayed in orange.}
    \label{fig:system}
\end{figure}

In this work, we present the first multilingual study across nine languages for speaker anonymization. 
Considering the potential of leaking personal information through latent channels in language-independent approaches, we choose to transform our previous language-dependent approach \cite{meyer2023prosody} into a multilingual one. This has the additional benefit of controllability in intermediate representations. 
The components in this system do not depend on specific models but only on high-level input representations which simplifies exchanging one model, e.g., for ASR, by a higher-performing multilingual one or a suite of monolingual models. This further allows to analyze the language dependence of each information channel (linguistic content, prosody, speaker) independently. For simple scalability to more languages, we use multilingual components instead of dynamically choosing a monolingual pipeline for each input. 
Further, we propose using Multilingual LibriSpeech (MLS) and CommonVoice (CV) to evaluate the privacy preservation across different languages, and present new trial subsets for them. 
We find that this multilingual speaker anonymization system is generally able to preserve the speakers' privacy in all languages of this study, to a similar extent as for English. This privacy protection, however, comes with a degradation of utility of the anonymized utterances for speech recognition. 
In an ablation study, we observe that this degradation is mainly due to the quality of the multilingual TTS model, and can likely be solved by using a better performing one. We conclude that it is not necessary to modify the actual anonymization part but only the language-dependent components in the system in order to open voice privacy technology to non-English-speaking communities.
All code, speaker verification trial files, and audio samples are available online\footnote{\url{https://github.com/DigitalPhonetics/speaker-anonymization}}.

\section{Multilingual Speaker Anonymization}

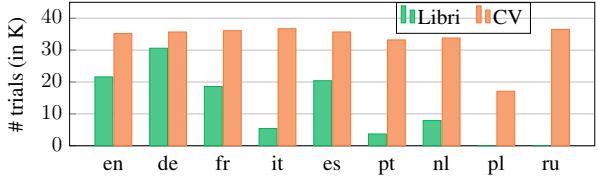
\begin{figure}
    \centering
    \resizebox{\columnwidth}{!}{
    \Large{
    \begin{tikzpicture}
        \begin{axis}[
            ylabel={\# trials (in K)},
            ybar=2*\pgflinewidth,
            symbolic x coords={en,de,fr,it,es,pt,nl,pl,ru},
            xtick = data,
            x tick label style=below,
            typeset ticklabels with strut,
            xticklabel shift={-5pt},
            ymin=0, ymax=45,
            ytick pos=left,
            ytick={0,10,20,30,40},
            ymajorgrids=true,
            major x tick style = transparent,
            bar width=12pt,
            width=14cm, 
            height=5cm,
            clip=false,
            legend style={at={(0.75,1.0)}, anchor=north, overlay, fill=none,
                /tikz/every even column/.append style={column sep=0.3cm}},
            legend columns=2
        ]

        \addplot[color=color1, fill=color1!70]
        coordinates {(en, 21.6)(de, 30.6)(fr, 18.6)(it, 5.4)(es, 20.4)(pt, 3.7)(nl, 7.9)(pl, 0)(ru, 0)};
        \addplot[color=color2, fill=color2!70]
        coordinates {(en, 35.2)(de, 35.7)(fr, 36.1)(it, 36.7)(es, 35.7)(pt, 33.2)(nl, 33.8)(pl, 17.1)(ru, 36.5)};
        \legend{Libri, CV}
        \end{axis}
    \end{tikzpicture}
    }}
    \caption{Number of trials per language and dataset, in thousand (K).}
    \label{fig:num_trials}
\end{figure}

\subsection{Monolingual Base System}
The speaker anonymization system has been proposed in \cite{meyer2023prosody}, as extension of \cite{meyer22b_interspeech, meyer2023anonymizing}, 
and is shown in Figure \ref{fig:system}. It consists of three phases. First, in the feature extraction, three types of information are extracted from the input speech: (a) The speaker embedding using a function based on Global Style Tokens (GST) \cite{gst} that is trained jointly with the TTS model. (b) The prosody of the speech in form of phone-wise pitch, energy, and duration values, with each value being normalized by the sequence mean to remove speaker-specific ranges \cite{exact}. (c) The linguistic content as phonetic transcription, recognized by an end-to-end ASR model based on a hybrid CTC/attention architecture. The second step is the anonymization, in which the original speaker embedding is replaced by an artificial one, generated by a Generative Adversarial Network, while using cosine distance to enforce a certain dissimilarity between original and anonymous speaker \cite{meyer2023anonymizing}.
Finally, the transcription, prosody, and anonymous speaker embedding are fed to a speech synthesis system consisting of a FastSpeech-2-like TTS~\cite{fastspeech2} and a HiFiGAN vocoder~\cite{hifigan}, using the IMS Toucan \cite{lux2021toucan} toolkit. 
 The output of this system is a waveform, with (ideally) the only difference between input and output being the speaker's voice.

\subsection{Multilingual Speech Recognition}
In the base system, the \asren\, model has been trained only on English data. As training a high-performing multilingual ASR is challenging, 
we simply replace \asren\, by the best performing freely available multilingual ASR model, which currently is Whisper large-v.3\footnote{\url{https://huggingface.co/openai/whisper-large-v3}} \cite{radford2023robust}. This model, \asranonwhisper\,, has been trained jointly on multiple tasks, including language identification and speech recognition, and supports 99 languages. 
The language of the input can either be given explicitly to the model or inferred by it, and is then forwarded to the TTS system.

\subsection{Multilingual Speech Synthesis} \label{sec:tts}
We replace the monolingual TTS in the base system by a multilingual one that is provided in the same toolkit\footnote{\url{https://github.com/DigitalPhonetics/IMS-Toucan/releases/tag/v2.5}}. This system has been proposed by \cite{lux2022low} and refined in \cite{lux2023toucan}. We use the provided off-the-shelf model and describe only the differences to the monolingual version. The system remains a FastSpeech-2-like synthesis \cite{fastspeech2} but is conditioned on language IDs. The input to the text encoder is enriched with language embeddings learned through a lookup table, while the spectrogram decoder is conditioned on the output of a jointly learned GST model to capture speaker characteristics. The GST model is modified with a style token disentanglement loss and 2000 style tokens \cite{Wu2022AdaSpeech4A}. Further, the speaker condition is reintroduced after every block in the decoder.  
The resulting spectrogram is passed to a HiFiGAN vocoder~\cite{hifigan}, similar to the one in the base system.  

\subsection{Model Statistics}
The total system consists of roughly 1.6 billion parameters for ASR (1.5B), speaker extraction (2M), speaker anonymization (50k), and speech synthesis (46M). 
The anonymization of one 10-second utterance takes 2 minutes on CPU (
Intel Xeon, E5-2687W v4), and 35 seconds on GPU (NVIDIA RTX 3090).

\begin{table}
    \centering\scriptsize
    \caption{Monolingual privacy and utility results for English, on original data and data anonymized by either the previous English-only \asren~ or the multilingual \asranonwhisper.}
    \resizebox{\columnwidth}{!}{
    \begin{tabular}{ll|cc|cc}
        \toprule
       & & \multicolumn{2}{c|}{Privacy - EER (\%) $\uparrow$} & \multicolumn{2}{c}{Utility - WER (\%) $\downarrow$}\\
        & &  \asvorig & \asvanon & \asrlibri & \asrwhisper\\
         \midrule
      \multirow{3}{*}{LS} & Original  & 4.59 & 11.43 & 3.07 & 2.8 \\
       & Anon-en & 46.82 & 34.29 & 7.26 & 6.26 \\ 
       & Anon-whisper & 47.50 & 34.77 & 8.04 & 6.67 \\
       \midrule
       \multirow{3}{*}{VCTK} & Original  & 2.37 & 13.68 & 10.95 & 1.58 \\
       & Anon-en & 43.44 & 40.00 & 16.92 & 9.46 \\
       & Anon-whisper & 44.22 & 39.92 & 20.05 & 12.28 \\
       \midrule
       \multirow{3}{*}{CV} & Original  & 5.08 & 12.44 & 28.23 & 6.80 \\
       & Anon-en & 41.45 & 29.79 & 38.34 & 29.91 \\
       & Anon-whisper & 40.38 & 29.56 & 34.40 & 21.35 \\
       \bottomrule
    \end{tabular}
    }
    \label{tab:english_results}
\end{table}

\section{Evaluation Plan}

\subsection{Data}

\begin{table*}
    \centering\scriptsize
    \caption{Multilingual privacy and utility results on original and anonymized speech.}
    \resizebox{\linewidth}{!}{
    \begin{tabular}{cc||rr|rrr||rr|rr}
        \toprule
        & & \multicolumn{5}{c||}{Privacy - EER (\%) $\uparrow$} & \multicolumn{4}{c}{Utility - WER (\%) $\downarrow$}\\
        & & \multicolumn{2}{c|}{Original speech} & \multicolumn{3}{c||}{Anon-whisper speech} & \multicolumn{2}{c|}{Original speech} & \multicolumn{2}{c}{Anon-whisper speech}\\
        dataset & lang & \asvorig & \asvvox & \asvorig & \asvanon & \asvvox & \asrwhisper & \asrmms & \asrwhisper & \asrmms \\
        \midrule
        \multirow{7}{*}{Libri} & en & 4.59 & 0.40 & 47.50 & 34.77 & 45.22 & 2.80 & 4.18 & 6.67 & 12.24\\
        & de & 1.63 & 0.16 & 46.55 & 24.94 & 46.70 & 5.45 & 16.09 & 13.54 & 25.85\\
        & fr & 0.43 & 0.15 & 47.75 & 29.16 & 46.16 & 5.27 & 21.02 & 16.82 & 32.83\\
        & it & 0.87 & 0.10 & 46.10 & 22.61 & 45.99 & 9.09 & 13.87 & 19.07 & 25.18\\
        & es & 0.14 & 0.20 & 41.90 & 29.89 & 42.46 & 4.02 & 15.51 & 10.00 & 24.10\\
        & pt & 0.41 & 0.00 & 40.53 & 14.16 & 40.28 & 8.50 & 22.94 & 21.41 & 44.01\\
        & nl & 0.08 & 0.00 & 45.41 & 24.42 & 35.99 & 10.07 & 12.54 & 21.50 & 28.69\\
        \midrule
        \multirow{9}{*}{CV} & en & 5.08 & 2.11 & 40.38 & 29.56 & 42.21 & 6.80 & 15.39 & 21.35 & 35.53\\
        & de & 2.97 & 1.69 & 42.67 & 34.72 & 47.20 & 4.07 & 18.19 & 14.11 & 34.30\\
        & fr & 4.12 & 3.84 & 48.04 & 43.06 & 47.50 & 12.10 & 29.63 & 25.52 & 45.60\\
        & it & 2.85 & 1.51 & 45.40 & 36.44 & 46.27 & 5.00 & 13.17 & 20.24 & 32.71\\
        & es & 5.75 & 2.68 & 45.90 & 38.54 & 45.78 & 5.03 & 19.23 & 22.71 & 38.74\\
        & pt & 5.62 & 3.26 & 47.38 & 41.64 & 46.68 & 6.21 & 27.97 & 30.68 & 61.50\\
        & nl & 3.75 & 1.66 & 39.57 & 38.37 & 42.24 & 4.55 & 13.14 & 26.33 & 45.61\\
        & pl & 4.18 & 3.39 & 42.77 & 43.31 & 46.00 & 6.06 & 45.13 & 28.09 & 60.88\\
        & ru & 4.61 & 3.45 & 48.21 & 39.81 & 46.39 & 6.05 & 101.75 & 42.24 & 103.73\\
         \bottomrule
    \end{tabular}
    }
    \label{tab:multiling}
\end{table*}

We perform the experiments on parts of MLS \cite{pratap20mls} for German (de), Italian (it), French (fr), Spanish (es), Portuguese (pt), and Dutch (nl), and of CV \cite{commonvoice2020} version 16.1\footnote{\url{https://commonvoice.mozilla.org/en/datasets}}, for the same languages plus English (en), Polish (pl), and Russian (ru). We further evaluate English on the test splits of LibriSpeech (LS)\cite{panayotov2015librispeech} and VCTK \cite{yamagishi2019vctk} as provided by the Voice Privacy Challenge (VPC) \cite{tomashenko2021voiceprivacy, tomashenko2022voiceprivacy}. 
For CV, we randomly select for each language and gender 10 speakers for enrollment and 5 additional speakers for trial. For Polish, having less audios by female speakers, we select 10 enrollment speakers per gender and 4 additional trial speakers. The test set of MLS consists of less speakers in total, so we use all available speakers for enrollment and trial, except in the case of German in which we use 10 speakers for enrollment and an additional 5 for trial. The lowest number of speakers in a language is 3 per gender for Dutch MLS. We restrict CV to 70 utterances per speaker, and select in both datasets 15\% of all utterances per speaker or at least 5 for enrollment, and the rest for trial. The total number of trials per language are shown in Figure \ref{fig:num_trials}. In the following, we refer to MLS and LS jointly as \textit{Libri}.
We publish all trial files with further details on Github to encourage their future use for speaker anonymization.

\subsection{Metrics}
We perform the evaluation as given in the protocol of the VPC 2022\footnote{The latest VPC 2024 has not been launched at time of submission.} \cite{tomashenko2022voiceprivacy}.  
Privacy is measured as Equal Error Rate (EER) of a speaker verification system (ASV) trying to identify the real speaker of each utterance. A failure of the ASV in form of random guesses signals successful anonymization, thus, we aim for an EER close to 50\%. 
We report 1-EER whenever the EER would be above 50 because the ASV could in such cases learn to flip its decisions to reach a better performance. Thus, a higher EER is better if the data is anonymized. 
We apply three ASV models of ECAPA-TDNN \cite{desplanques2020ecapa} architecture with cosine distance, which only differ in their training data: Following the VPC 2022, \asvorig\, and \asvanon\, are trained on original and anonymized versions of LS train-clean-360. \asvvox\, is trained on the original VoxCeleb 1 and 2 datasets \cite{Nagrani17voxceleb,Chung18bvoxceleb} by SpeechBrain \cite{speechbrain}\footnote{\url{https://huggingface.co/speechbrain/spkrec-ecapa-voxceleb}}. 
Test data is anonymized on speaker-level by keeping the same target for each input speaker in a session, whereas training data is anonymized on utterance-level. 
We report the average score of the female and male ASV trials.

Utility is evaluated by comparing the word error rates (WER) of performing speech recognition on anonymized and original data. We apply three ASR models: \asrlibri\, which is a transformer+CTC model with transformer language model, trained on LS train-clean-360, and provided in the VoicePAT toolkit \cite{meyer2024voicepat}. \asrwhisper\, is the same Whisper model as used in the multilingual anonymization. For comparison, we apply the large multilingual \asrmms\, for 1162 languages\footnote{\url{https://huggingface.co/facebook/mms-1b-all}} of the MMS project \cite{pratap2023mms}. Lower WER scores correspond to higher utility.

\section{Experiments}

\subsection{English Anonymization using Whisper}

Before testing the system on multilingual data, we first analyze the impact of the proposed changes on the anonymization of English speech, using either the monolingual \asren\, or the multilingual \asranonwhisper\, (Table \ref{tab:english_results}). 
In terms of privacy, the choice of ASR model does not have much effect. This is different for utility, in which using \asren~ leads to lower WER results for LS and VCTK since it has been optimized for this data. For the more general CV, using \asranonwhisper\, during anonymization seems to be the better choice. 
As an evaluation model, \asrwhisper\, clearly outperforms the previous \asrlibri\,. This is most obvious when comparing the WER scores of both models on the original data of VCTK and CV but can be also observed for the anonymized data, regardless of whether Whisper has been part of the anonymization or not. 
It should be noted that the multilingual TTS has a lower quality than the previous monolingual TTS, which achieved WER scores of 4.62 on LS and 10.63 on VCTK with \asrlibri. 
The results generally show that CV is a more challenging dataset for voice privacy than LS and VCTK, indicated by lower privacy scores and higher utility losses. While still being a single-speaker dataset under relatively controlled conditions, this dataset is arguably a better proxy for judging how well different anonymization system would perform in a real-world application.

\subsection{Multilingual Anonymization}

\begin{table*}
    \centering\scriptsize
    \caption{Ablation results when disabling the anonymization (resys and gold-resys) and ASR components (gold-resys and gold-anon).}
    \resizebox{\linewidth}{!}{
    \begin{tabular}{cc|rrrrr|rrrrr|rrr}
        \toprule
        & & \multicolumn{5}{c|}{EER (\%) $\uparrow$} & \multicolumn{5}{c|}{WER (\%) $\downarrow$} & \multicolumn{3}{c}{PER (\%) $\downarrow$}\\
        data & lang & original & anon & resys & gold-resys & gold-anon & original & anon & resys & gold-resys & gold-anon & original & anon & gold-resys\\
        \midrule
        \multirow{3}{*}{Libri} & en & 4.59 & 47.50 & 26.94 & 26.27 & 42.89 & 2.80 & 6.67 & 5.85 & 6.09 & 7.81 & 1.00 & 3.92 & 3.64\\
        & de & 1.63 & 46.55 & 12.12 & 12.65 & 46.75 & 5.45 & 13.54 & 12.31 & 12.96 & 12.48 & 2.41 & 6.60 & 5.55 \\
        & it & 0.87 & 46.10 & 9.74 & 9.49 & 46.76 & 9.09 & 19.07 & 20.30 & 15.85 & 17.11 & 2.66 & 7.62 & 5.99 \\
        \midrule
        \multirow{3}{*}{CV} & en & 5.08 & 40.38 & 25.39 & 25.12 & 43.63 & 6.80 & 21.35 & 17.95 & 16.24 & 19.69 & 3.26 & 13.91 & 10.71\\
        & de & 2.97 & 42.67 & 24.82 & 24.47 & 45.22 & 4.07 & 14.11 & 12.70 & 11.64 & 13.75 & 1.74 & 7.41 & 6.04\\
        & it & 2.85 & 45.40 & 23.07 & 23.46 & 45.36 & 5.00 & 20.24 & 19.65 & 17.74 & 19.24 & 1.56 & 11.72 & 10.56\\
        & pl & 4.18 & 42.77 & 30.82 & 31.64 & 42.10 & 6.06 & 28.09 & 21.95 & 18.16 & 24.34 & 1.99 & 11.54 & 7.92 \\
        \bottomrule
    \end{tabular}
    }
    \label{tab:analysis}
\end{table*}

\subsubsection{Privacy according to English ASV}
The privacy results for the multilingual datasets are shown in the left columns of Table \ref{tab:multiling}. Although \asvorig\, and \asvanon\, have been trained only on English speech, they perform in most cases better on the original speech of languages other than English. On anonymized data, the EER scores of \asvorig\, indicate generally a similar level of privacy for all languages. However, when using \asvanon, the privacy scores for non-English speech are lower on LS but higher on CV. It is possible that a language-specific model trained on anonymized data of that language would perform better even on CV. However, based our results, we assume that a speaker verification attacker in a real-world scenario that expects anonymized English speech would fail to identify the anonymized speakers of other languages. 

\subsubsection{Privacy according to multilingual ASV}
In order to test the influence of the language mismatch between ASV training and evaluation, we further compute the EER when using \asvvox\, which has been trained on the original data of the monolingual VoxCeleb 1 and the multilingual VoxCeleb 2. On original data, all scores by \asvvox\, are consistently better than with the English \asvorig, for LS already close to perfect. For CV, the relation between the EER scores of all languages is similar to the ones using \asvorig, thus, the multilingual ASV performs well on the same languages as the monolingual one. On anonymized speech, the results of both ASV models are similar, except for the Dutch LS data. We also perform these experiments with multilingual ASV models trained on the train set of MLS and observe similar trends, but overall worse scores, especially on original data. This suggests that the main reason why \asvvox\, achieves lower EER on the original data is actually due to it being trained on more data than \asvorig\, which seems to be more important than this data being multilingual.

\subsubsection{Utility}
In terms of utility, the differences between the languages are more obvious. For most languages and datasets, the WER as measured by \asrwhisper\, is twice to thrice as high on the anonymized data than on their original counterparts. For some languages, however, the WER increases more drastically. For example, Russian increases by a factor of 7 from 6.05 to 42.24. This degradation could result to anonymized data being less usable than non-anonymized ones in an application. 
Since Whisper is used during both anonymization and evaluation, it is possible that this introduces a bias and that the actual intelligibility loss would be higher when using a different ASR during evaluation. Therefore, we additionally test the speech with \asrmms\, as second multilingual ASR. However, already the WER scores on the original data turn out to be substantially worse than the results of \asrwhisper.  
On anonymized data, for most datasets, the WER according to \asrmms\, increases less than according to  \asrwhisper, i.e., only up to two times as much. We therefore conclude that using the same ASR in both anonymization and evaluation is not likely to introduce a bias but that it is important to choose a strong ASR model for evaluation to be able to judge the utility decrease accurately.

\section{Analysis} \label{subsec:analysis}

In order to understand the influence of each component on the results, we perform an ablation study in which we disable either the anonymization (\textit{resynthesis}), ASR transcription (\textit{gold-anonymization}), or both (\textit{gold-resynthesis}). In all experiments, the prosody extraction and the speech synthesis remain unchanged. Due to space limitations, Table \ref{tab:analysis} shows the results only for a selection of languages but the same trends hold for the languages not shown.

\subsection{Impact of the ASR Module}
In the first experiment, we compare the effect of using either gold transcripts or the ASR module to extract the linguistic content in the input speech. The privacy and utility scores are shown in the columns \textit{gold-anon} and \textit{anon}, respectively. Both metrics reveal only little influence of the speech recognition during anonymization. Naturally, using gold transcripts leads to a lower WER than using recognized transcripts but the difference is in most cases negligible. 

\subsection{Impact of the Anonymization Module}
Next, we disable the anonymization module by instead using the speaker embedding of the original input speaker during synthesis. These results can be seen in the \textit{resys} columns. Comparing this to \textit{anon} shows again only little effect on the utility scores. However, as expected, using anonymous speaker embeddings greatly improves the privacy results, leading to a difference of EER as large as 9.49 to 46.76 in Italian MLS.

\subsection{Impact of Speech Synthesis}
Since the utility difference between original and anonymized data could not be explained by the inclusion of neither speech recognition nor anonymization, we finally test disabling both, and use the gold transcript and original speaker embedding for resynthesis (\textit{gold-resys}). The only difference between this condition and the \textit{original} results is the use of the speech synthesis model. This reveals that the synthesis has by far the most influence on utility scores, and also a large impact on the privacy. The effect is most obvious for Polish CV, in which the synthesis alone increases the EER from 4.18\% to 31.64\% and the WER from 6.06\% to 18.16\%. Qualitative analysis reveals that this degradation is mainly due to a decrease in speech quality rather than due to synthesis errors or an actual intelligibility loss. We therefore compute the Phone Error Rate (PER) using the phonemization in IMS Toucan. As shown in the right-most block of Table \ref{tab:analysis}, these error rates are generally much lower than for full words. The degradation through synthesis is still observable but the general intelligibility of the synthesized speech seems to be higher than expected according to the WER. Overall, the results of the ablation study suggest that replacing the speech synthesis system used in this work with a stronger model or speech enhancement on top of the output might decrease the strong impact of the synthesis and significantly improve the outcome.

\section{Conclusion}
In this paper, we present the first multilingual study of speaker anonymization across nine languages. We propose changes to a state-of-the-art controllable but language-dependent system in order to transform it into a multilingual anonymization tool. To test multilingual voice privacy tools, we present new verification trial splits for Multilingual LibriSpeech and CommonVoice. Experiments on this data show a general suitability of this system to preserve speaker privacy but at the cost of speech quality degradation which reduces the utility of the anonymized speech for downstream applications. However, an ablation study revealed that this degradation is mainly due to the performance of the speech synthesis module and could be improved by replacing this component or applying speech restoration to its output. The conclusions of this work are limited by using only common Indo-European languages and only monolingual audios. 
In summary, however, this work shows that it is possible to make speaker anonymization systems applicable to several languages while maintaining certain monolingual components, alleviating the complexity of integrating new languages without suitable task specific data. This is an important step towards voice privacy beyond a selected few languages.

\section{Acknowledgements}
This work is funded by the Deutsche Forschungsgemeinschaft (DFG, German Research  Foundation) – Project: Multilingual Controllable Voice Privacy (VoiPy) - Project number 533241795.

\bibliographystyle{IEEEtran}
\bibliography{mybib}

\end{document}